# Decision-Making Technology for Autonomous Vehicles: Learning-Based Methods, Applications and Future Outlook

Qi Liu, Xueyuan Li, Shihua Yuan, Zirui Li

*Abstract*— Autonomous vehicles have a great potential in the application of both civil and military fields, and have become the focus of research with the rapid development of science and economy. This article proposes a brief review on learning-based decision-making technology for autonomous vehicles since it is significant for safer and efficient performance of autonomous vehicles. Firstly, the basic outline of decision-making technology is provided. Secondly, related works about learning-based decision-making methods for autonomous vehicles are mainly reviewed with the comparison to classical decision-making methods. In addition, applications of decision-making methods in existing autonomous vehicles are summarized. Finally, promising research topics in the future study of decision-making technology for autonomous vehicles are prospected.

## I. INTRODUCTION

With the development of science and economy, autonomous driving technology has gradually become the focus of researches since its broad application prospects in the field of both military and civilian. In the field of military, highly intelligent autonomous military vehicles can effectively assist soldiers in numerous tasks such as intelligence acquisition, fire strikes, and monitoring [1]; while in the field of civil, autonomous vehicles have a great potential in reducing traffic accidents and alleviating traffic congestion [2].

Autonomous vehicle is a comprehensive intelligent system that integrates environmental perception, path planning, decision-making and motion controlling technologies [3]. As the "brain" of autonomous vehicles, decision-making system is significant for the safe and efficient driving of vehicles, and how to design high intelligent and reliable decision system gradually become the focus of research in the field of autonomous driving. Decision-making is expressed to generate human-level safe and reasonable driving behaviors considering surrounding environmental information, motion of other traffic participants and state estimation of ego vehicles; then the generated driving behaviors are taken into account by the motion control system to achieve efficient operation of autonomous vehicle [4, 5].

In general, decision-making methods can be divided into classical methods and learning-based methods. Usually, autonomous vehicles are operating in complex, dynamic environment with the cooperation with other traffic participants. Classical methods can't be always effective in such driving environment due to poor robust, thus learning-based methods are utilized to achieve better decision-making for autonomous vehicles [6]; in addition, with the emergency of new powerful computational technologies, learning-based approaches have gained huge popularity and development in the field of autonomous vehicles [7].

There are several published reviews on decision-making for autonomous vehicles. A brief review about framework and typical methods for decision-making was presented in [8], while a survey on classical methods of fusion of planning and decision-making was summarized in [9]. In addition, a comprehensive overview of autonomous driving technology was summarized in [10, 11], however the content of decision-making was not detailed enough. Considering the importance of decision-making technology for autonomous vehicles and the tendency of recent development for related methods, the main difference in this article is presenting a systematic review mainly on learning-based decision-making methods that have emerged in recent years with the summarization of applications in existing autonomous vehicles. We hope that this review study will contribute to the research of decision-making methods for autonomous vehicles in the future.

This article is organized as follows. Section II presents a detailed introduction of the general outline and framework for decision-making system. Section III reviews learning-based methods related to decision making for autonomous vehicles as well as some respectable related works using classic methods. Section IV summarizes the applications of decision-making methods carried out in existing vehicles. Section V discusses the future development direction and research focus of decision-making methods for autonomous vehicles.

## II. THE OUTLINE OF DECISION-MAKING SYSTEM

In order to carry out specific research for decision-making, a clear understanding of the general framework of decision-making system in autonomous driving technology is essential for designing efficient methods. This section presents a general outline of decision-making system in autonomous vehicles based on the summary of related researches. Four aspects: specifically, inputs and outputs (IOs), design criteria, design constrains and applications scenarios of decision-making system for autonomous vehicles are summarized in the following contents. In addition, the complete designing framework of decision-making system is illustrated in Figure 1.

### A. Inputs and Outputs of Decision-Making System

Decision-making system in autonomous vehicles is the transition of environmental perception system and motion

All authors are with the School of Mechanical Engineering, Beijing Institute of Technology, Beijing, China. (E-mails: 3120195257@bit.edu.cn; lixueyuan@bit.edu.cn; yuanshihua@bit.edu.cn; 3120195255@bit.edu.cn ).

Zirui Li is also with Department of Transport and Planning, Faculty of Civil Engineering and Geosciences, Delft University of Technology, Stevinweg 1, 2628 CN Delft, The Netherlands.

(Corresponding author: Xueyuan Li and Zirui Li)

planning system. In general, the inputs of decision-making system are environmental clues and status of ego vehicle, while the outputs are a serious of strategies including high-level behaviors and low-level control commands that are fed into motion planning system [12].

Specifically, the inputs of decision-making system can be summarized as following aspects:

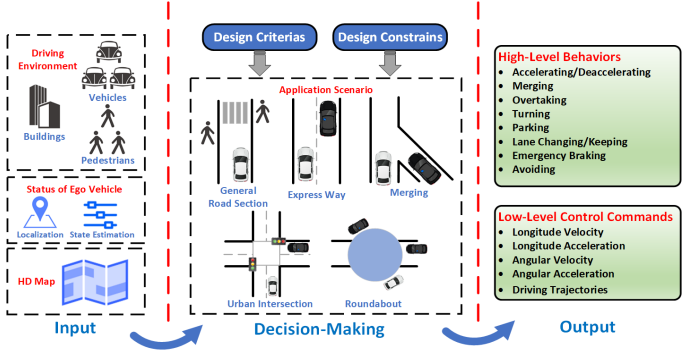

Figure 1. Designing framework of decision-making system

*Surrounding environmental information.* Raw data is usually collected from different types of sensors (Lidar, camera, radar, etc.) equipped on vehicles, and then are processed to generate perception results mainly including static and dynamic objects information, road information and traffic signs information.

*Status of ego vehicles.* It mainly represents location acquired by GNSS/IMU system and motion information from motion estimation system.

*High-Definition Map (HD Map).* A wealth of information accurate to the lane level can be provided by HD Map, that can be utilized as an auxiliary means of environmental perception system of ego vehicles to enhance the perception accuracy and reduce the computational cost.

The outputs of decision-making system can be concluded as below:

*High-level behaviors,* such as merging, overtaking, lane keeping and lane changing.

*Low-level control commands,* mainly including longitude velocity, acceleration and angular velocity, acceleration.

### B. The Design Criteria of Decision-Making System

The purpose of decision-making system is to generate human-like safe and high reliable driving strategy, a serious of design criteria are needed to be formulated to achieve better decision and five aspects are summarized below [10, 13]. Good real-time performance for decision-making; balance between driving safety and efficiency (usually the priority of safety is higher than driving efficiency); reasonable and correct generated decision; ride comfort of vehicles (steering stability, less emergency brake); high capability of faults detection.

### C. The Design Constrains of Decision-Making System

Researches of decision-making methods needs to consider numerous types of factors to achieve a more complete system; several designing constrains for decision-making system can be extracted from related works that are listed as follows.

*Information of surrounding environment.* In general, objects information within a certain distance around ego vehicle need to be considered. For instance, position and speed of other vehicles; the prediction of pedestrian and vehicles behaviors; static obstacles placed or dropped on roads; road drivable areas; traffic and road signs.

*Information of local traffic regulations.* This constrain mainly refers to the following of traffic rules for ego vehicles when making decision, including road speed limits, U-turn allowance, no parking, etc.

*Current status of ego vehicles.* This part includes the location, speed and heading of ego vehicles; the current lane, and the next lane to be entered should also be considered.

*Results of path planning.* Path planning can be divided into global path planning and local path planning, and decision-making process mainly considers the results of local path planning in the current environment.

*Historical decision-making results.* This part specifically refers to the sequence of historical decisions made by the ego vehicles at the last moment (or the previous few moments), which should be taken into account for decision-making at the current moment.

*Driving ethics.* This part refers that vehicles must comply with driving ethics during operation [14], such as giving courtesy to pedestrians, giving ways to special vehicles (ambulances, firetruck), turning off the high beams for opposite vehicles at night, etc.

### D. Application Scenarios of Decision-Making System

Decision-making is required in almost all scenarios, as long as autonomous vehicle is in operation status. With the increasing requirements for decision-making systems due to the complexity of driving environment, related researches are focusing on V2V or V2P cooperation in some typical scenarios including general road section, express way, urban intersection, merging traffic and roundabout.

### III. REVIEW OF DECISION-MAKING METHODS

In this section, learning-based decision-making methods for autonomous vehicles are mainly overviewed; in addition, related works using classical methods for decision-making are also summarized for the completeness of different methods. Characteristic of different methods are compared in TABLE I.

### A. Classical Methods

In general, classical methods for decision-making can be divided into three categories: rule-based methods, optimization methods and probabilistic methods.

#### 1) Rule-Based Methods

Rule-based decision-making methods depend on a rule database constructed according to numerous traffic laws, driving experience and driving knowledge; strategies are then determined considering different status of vehicles.

The most representative of rule-based methods is the Finite State Machine (FSM) methods. FSM is a mathematics model with discrete input and output, corresponding actions are generated depending on the responding to external events and states of agents are then transited from one to another. FSM can be divided into three categories according to the logic structure of different states, specifically tandem type, parallel type and hybrid type [13]. Characteristic of different types of FSM are illustrated in TABLE II.

TABLE I. CHARACTERISTIC OF DIFFERENT METHODS FOR DECISION-MAKING

| Methods | | Refs | Pros | Cons |
|---|---|---|---|---|
| Classical Methods | Rule-based Methods | [12-18] | • Strong interpretability and adjustability<br>• Strong feasibility of implementation since its low requirements for hardware<br>• Good decision-making breadth | • Difficult to handle complex driving conditions since the lack of decision-making depth<br>• Poor robustness for dynamic driving environment |
| | Optimization Methods | [19-26] | • Optimized decisions can be generated<br>• Interaction between different traffic participants can be better modeled | • The assumption of "optimal strategy" for agents is often inconsistent with practical applications |
| | Probabilistic Methods | [27-29] | • Convenient to combine with other types of methods | • Low computational efficiency and difficult to generate optimal decision in complex environment |
| Learning-Based Methods | Statistic Learning-Based Methods | [30-32] | • Good versatility<br>• Suitable for simple scenarios with sufficient environmental information | • Requirement for plenty of training datasets<br>• Low decision-making accuracy |
| | Deep Learning-Based Methods | [33-39] | • High decision-making accuracy for specific scenarios<br>• End-to-end system ensures the full utilize of environmental information | • Poor universality of algorithms in dynamic scenario.<br>• Requirement for plenty of training datasets thus quality of the datasets will greatly influence the effect of algorithm |
| | Reinforcement Learning-Based Methods | [11, 40-54] | • Better modeling of uncertain and dynamic environments<br>• Flexible framework of algorithms with high expandability | • Greatly depends on the establish of reward function<br>• Poor stability, over-fitting in DRL methods. |

TABLE II. COMPARISON OF DIFFERENT TYPES OF FSM

| Types | Characteristic | | |
|---|---|---|---|
| | Features | Pros | Cons |
| Tandem | Sub-states are connected in tandem structure with unidirectional state transmission. | • Good traversal depth<br>• Suitable for simple decision scenarios | • Poor system stability |
| Parallel | Sub-states are arranged in multi-point connection structure to provide parallel decision. | • Good traversal breadth<br>• Parallel processing of multiple states<br>• High system stability and expandability | • High system complexity<br>• Lack of traversal depth |
| Hybrid | Sub-states are connected in both tandem and parallel structure. | • Good traversal depth and breadth.<br>• Wide range of application scenarios | • High system complexity<br>• Low computational efficiency |

FSM methods have already been widely used in some existing typical autonomous vehicles. For instance, tandem type was utilized in Talos [15]; parallel type was used in Junior [16] and Bertha [17]; while hybrid type was carried out in Odin [18].

Apart from typical FSM that has been utilized in existing vehicles, some researches are also carried out depending on a specific rule base. In [19], a hybrid flow diagram was designed for generating decision results through a series of judgment conditions of the surrounding environment; the feasible trajectory was selected without detailed prior maps and algorithms were verified on "AUTOPIA" instrument vehicle. In [20], a "multi-point turn" decision framework was proposed for vehicle steering control basing on a rule that minimized steering widths. In [21], a hierarchical framework was established for tactical and strategical behavior generation; the framework had the advantage of high generality and expandability that can be combined with various of scenarios.

*2) Optimization Methods*

Optimization methods usually rely on a reward or utility function to generate decision results.

Model Predictive Control (MPC) is one of the feasible methods for decision-making. In [22], MPC was carried out for the controlling of whole traffic situation, where high-level behaviors of ego vehicle were generated and other vehicles were indirectly influenced by the ego's behaviors. In [23], vehicle-pedestrian interaction was considered through a multi-state forced pedestrian motion prediction model, MPC was then utilized to generate low-level control commands for ego vehicle. In [24], MPC was combined with Inverse reinforcement learning (IRL) to establish more suitable cost function.

Game-Theory is another optimization method for solving decision-making problems. An assumption that all agents will adopt the "optimal strategy" is put forward first and behaviors are generated according to the corresponding strategies of other agents. In [25], a Stackelberg game was utilized to solve the decision-making for more than 30 vehicles; however, it suffered from low average speed and large number of constraint violations [8]. In [26], market-based mechanisms were established for modeling the cooperative behavior of platoons of autonomous vehicles; two auction clearing rules were tested in a microscopic traffic model and showed a stable cooperation. In [27], an interactive game tree was designed for cooperation of vehicles in the merging scenario of dense urban traffic; results showed a good real-time performance with 0.08s for the entire decision-making loop.

Evaluation-based methods can also be utilized for selecting optimum driving behaviors. In [28], a multiple attribute-based decision-making method was proposed based on AHP and TOPSIS; AHP was used for obtaining the weights of different attributes, while TOPSIS was used to generate optimum behaviors. The algorithm had been verified on "Intelligent Pioneer" platform in urban scenario.

*3) Probabilistic Methods*

Probabilistic methods generate behavior results basing on the probability theory in mathematics. A probabilistic model needs to be established for determination of behaviors.

In [29], a probabilistic graphical model (PGM) was carried out to estimate intentions of surrounding vehicles in the merging scenario, and motion commands were generated through an off-the-shelf ACC distance keeping model without the requirement of acceleration information of other vehicles. In [30], a robust Two-Sequential Level Bayesian Decision Network (TSLDN) was proposed for decision-making in lane change scenario; risk assessment through an Extended collision (ETTC) and Dynamic Predicted Inter-Distance Profile (DPIDP) were developed to ensure safe probabilistic behaviors generation. In [31], a similar work was carried out by authors from [30]; the original decision framework was extended to handle evasive action selection and Extended Kalman Filter was also combined to ensure more safer behaviors.

*B. Learning-Based Methods*

Learning-based methods refers to the utilization of artificial intelligence technologies to achieve decision-making for autonomous vehicles. Usually, driving data samples need to be established first, and different learning methods or network structures are then adapted to realize autonomous learning of vehicles to generate reasonable decisions based on different environmental information. Learning-based methods can be divided into three categories: statistic learning-based methods, deep learning-based methods and reinforcement learning-based methods. In addition, some datasets [32, 33] can also be used for the verification of learning-based decision-making algorithms for autonomous vehicles.

*1) Statistic Learning-Based Methods*

Statistic learning-based methods enable autonomous vehicles to master human-like decision-making capabilities through a large amount of training data. Typical statistic learning-based methods include SVM, AdaBoost, etc.

In [34], SVM was trained for decision-making in lane change scenario taking relative position and velocity as inputs; trajectory was generated with the combination of MPC with several safety constraints. On the basis of [34], SVM was also carried out in [35] for lane change decision-making; more constraints including lane change benefits, safety and tolerance were considered, and Bayesian parameters optimization was adopted for better determination of parameters in lane change model; the algorithm was finally verified in "Zhongtong" autonomous bus with an accuracy of 86.27%.

Apart from SVM methods. In [36], AdaBoost was used for decision-making in "Cut-In" scenario with risk assessment; speed of ego vehicles and the agents were chosen as input while distance to collision (DTC) was chosen as output. Results showed that this method can fulfill safe manoeuvre.

*2) Deep Learning-Based Methods*

The framework of deep learning method is similar to that of traditional machine learning, the main difference is that deep learning methods utilize neural network structure to learn the features of data, and generate classification or regression results. End-to-end systems have been designed to achieve decision-making for autonomous vehicles thanks to the advantages of deep learning methods in image processing. Usually, sensors data are chosen as input and low-level control commands are generated through the trained neural networks.

Vision sensor is widely equipped on autonomous vehicles because of its high performance-cost ratio, thus some end-to-end researches are developed based on the process of visual image. In [37], images captured from a single front-camera were chosen as input; a CNN was trained for end-to-end decision-making to generate steering commands for lane keeping. A similar work was carried out by NIVIDIA in [38], the main difference was that images from three cameras are used as input to generate steering commands. In [39], "DriveNet" was proposed based on the structure of CNN; three consecutive frames of images captured from a single front-camera were selected as input, and algorithm was verified in driving environments with various illumination condition.

To generate more complete control instructions, both straight and steering commands were computed through three types of inputs including visual images, measurement and high-level commands. In [40], two networks are trained, one combined all aforementioned factors together while another one considered high-level commands as a branched switch. Results showed that the latter performed better in accuracy of decision. In [41], Attention Branch Network (ABN) was designed to achieve end-to-end decision-making. Original visual images were input first, and the "attention map" was generated in the middle layer of the designed network. Finally, throttle and steering commands were generated by the combination of "attention map" and convolution feature of the original images with expected velocity of ego vehicle.

In addition to visual images, some researches were also carried out based on Lidar point cloud data (PCD). In [42], driving path was produced by FCN integrating Lidar PCD, IMU and navigation information from Google map. In [43], Lidar PCD was converted to a grid map first, then FCN and Inverse Reinforcement Learning (IRL) were combined to construct an end-to-end system with better cost function, and finally a serious of actions were generated. The above several exemplary works are summarized in TABLE III.

TABLE III. SUMMARY OF EXEMPLARY RELATED WORKS WITH DEEP LEARNING-BASED METHODS

| Refs | Network | IOs | | Performance | Hardware |
|---|---|---|---|---|---|
| | | Inputs | Outputs | | |
| [38] | CNN | Visual image | Steering command | 3 hours and 100 miles in total of driving in Monmouth County, NJ with 98% time of autonomours driving | NVIDIA DRIVE PX |
| [39] | DriveNet | Visual image | Steering commands | 15.81 RMSE of trajectory with 1.70ms/fps in 11.8km of driving datasets with different illumination condition | NVIDIA GTX1070 |
| [40] | CNN | Visual image | Straight and steering commands | 88% success rate in CARLA simulator; 2 hours driving nearly without missed turns in physical world | NVIDIA TX2 |
| [41] | ABN | Visual image and expected velocity | Throttle and steering commands | Defined the "autonomy score" to evaluate the algorithms: 92.7 for throttle control; 97.2 for steering control | Not mentioned |
| [42] | FCN | Lidar point clouds | Driving trajectories | 88.56% of precision in regions of 60×60 meters; 93.44% of precision in regions of 40×40 meters | NVIDIA GTX980Ti |
| [43] | FCN with IRL | Lidar point clouds | Discrete set of actions | Totally 120km of driving in a modified GEM golf cart | Not mentioned |

*3) Reinforcement Learning-Based Methods*

Reinforcement learning (RL) method is currently one of the most commonly used learning-based methods for decision-making. The goal of reinforcement learning is to learn strategies to maximize returns by trying various behaviors, behaviors of agents can be adjusted according to reward functions. Existing data as well as new data obtained through exploration of the environment can be used to update and iterate the existing model cyclically.

RL method usually depends on Markov Decision Processes (MDPs) to describe the interaction states of agent and environment. Considering that states cannot always be observed, Partially Observable Markov Decision Process (POMDP) has been proposed to describe the state space in a more realistic way. The key to decision-making with RL method is to efficiently solve MDPs or POMDP. Related approaches can be divided into typical solvers and solvers combining with deep learning methods.

*a) Solvers with Typical Methods*

In [44], interaction of pedestrians with vehicles was modeled as MDPs; two Q-functions for different scenario were combined to generate decision results. In [14], driving ethics were considered including three different policies, specifically ralwsian contractarianism, utilitarianism and egalitarianism; MDPs was proposed and solved based on typical Bellman's equation.

Apart from solving decision-making with MDPs, POMDP were also utilized by numerous researchers. In [45], POMDP was presented for decision-making through occluded intersection, and then solved by traversing a typical belief search tree. A similar method was proposed in [46], however the POMDP framework was solved by Adaptive Belief Tree (ABT) to achieve higher computing efficiency. Moreover in [47], POMDP was solved by "TAPIR" toolkit; and blind areas of ego vehicle were taken into account to generate safer behaviors.

It is vital to model the uncertainty of traffic environment when utilizing POMDP for decision-making. In [48], discrete Bayesian Network was carried out for the modeling of uncertainty of traffic environment; and POMDP was solved by fusion of "MCVI" and "SARSOP". While in [49], dangerous traffic situations were additionally considered and modeled by Dynamic Bayesian Network (DBN) to generate human-like behaviors with anticipation.

*b) Solvers Combining with Deep Learning Methods*

With the rapid development of deep learning methods in the field of supervised learning, deep reinforcement learning (DRL) methods have showed a large potential in high intelligent decision-making. The main idea of DRL is to incorporate neural networks into RL frameworks. Representative DRL methods including Deep Q Network (DQN), Deep Deterministic Policy Gradient (DDPG), and Asynchronous Advantage Actor-Critic (A3C).

Solving MDPs and POMDP with higher real-time performance is critical to their practical applications in autonomous vehicles. In [50], decision-making at intersections was modeled as Hierarchical Options MDP (HOMDP) that only considered current observation to reduce computational cost; algorithm was designed based on POMDP and solved by typical DQN. While in [51], research work focused on the improvement of DQN to improve computational efficiency; a "Rainbow DQN" with safe constraint was carried out to generate safe driving behaviors with high sample efficiency.

Learning to generate safer behaviors is also crucial in DRL framework. In [52], POMDP was utilized with the combination of "risk-sensitive" approach; the framework was solved by Offline Deep Distributional Q-Learning with Online Risk Assessment to achieve safe decision-making. In [53], decision-making at occluded intersection was modeled as MDPs; a stricter risk-based reward function was designed to punish risk situation instead of only collision happening; finally, a generic "risk-aware DQN" was proposed for solving the model. Moreover in [54], driving scenario was also modeled as MDPs; multiple neural networks were assembled with additional randomized prior functions to optimize the capacity of typical DQN in solving uncertain environments; the proposed algorithm has been verified to realize safe decision-making in a highly uncertain environment for autonomous driving.

Establishing a benchmark of different scenarios is very important for the verification of the DRL algorithms. In [55], POMDP was solved to generate collision-free behaviors by a hybrid algorithm named "HyLEAP" combining belief tree methods and neural networks; in addition, "OpenDS-CTS" benchmark was established for verification of decision-making mainly in car-pedestrian accident scenarios.

More types of tasks at intersection were considered in [56], the decision problem was modeled as MDPs and solved by "Multi-Task DQN" that has shown good performance in simulation of different intersection scenarios; experiment was also carried out in an unsignalized T-junction. In [57], different driving tasks were modeled through a hierarchical framework integrating both high-level policy and low-level control; high-level policy was generated by solving POMDP with Advantage Actor-Critic (A2C), while low-level control was computed through vehicle kinematic model. The above several exemplary works are summarized in TABLE IV.

TABLE IV. SUMMARY OF EXEMPLARY RELATED WORKS WITH REINFORCEMENT LEARNING-BASED METHODS

| Refs | Decision-Making | | | Scenario | Interaction | Verification |
|------|------|--------|-------|----------|-------------|--------------|
|      | Model | Solver | Level |          |             |              |
| [14] | MDPs | Search-based | Low | Straight road considering ethics | V2V | Simulation |
| [44] | MDPs | Defined Q-functions | Low | Straight road near school | V2P | Simulation |
| [47] | POMDP | TAPIR Toolkit | Low | Intersection considering blind spots | V2V | Simulation |
| [48] | POMDP | MCVI and SARSOP | High | Intersection without traffic signs | V2V | Simulation |
| [49] | POMDP | Search-based | Low | Intersection considering traffic flow forecast | V2V | Simulation |
| [50] | HOMDP | DQN | High | Typical Intersection | V2V | Simulation |
| [51] | MDPs | Rainbow DQN | High | Lane change in express way | V2V | Simulation |
| [53] | MDPs | risk-aware DQN | High | Occluded intersection | V2V | Simulation |
| [54] | MDPs | Ensemble RPF | High | Typical Intersection | V2V | Simulation |
| [55] | POMDP | HyLEAP | Low | Typical Intersection | V2P | Simulation |
| [56] | MDPs | Multi-Task DQN | High | Intersection without traffic signs | V2V | Simulation and experiment |
| [57] | POMDP | A2C | High; Low | Merging in express way | V2V | Simulation |

IV. APPLICATIONS OF DECISION-MAKING TECHNOLOGIES IN EXISTING AUTONOMOUS VEHICLES

This section summarizes the applications of decision-making methods in some typical existing autonomous vehicles as well as recent works with verification in real vehicles, that are arranged in TABLE V.

V. FUTURE OUTLOOK

In this section, the future research emphasis of decision-making for autonomous vehicles are forecasted on the following aspects.

*A. The Vehicle-Pedestrian Interaction*

The design of algorithms for decision-making algorithm must consider the interaction with other traffic participants. However, most research works focus on the decision-making under vehicle-vehicle interaction, and rarely consider vehicle-pedestrian interaction. Decision-making in view of interaction with pedestrians is crucial for safe driving since pedestrians are vulnerable traffic participants. Dividing the behavior of pedestrians with clear boundary, analyzing and predicting of actions for pedestrians should be integrated into the future researches for decision-making system.

*B. Safer and More Comfortable Decision-Making System*

Safety is the most important factor for driving, part of the research works has incorporated safety into the consideration of decision-making [58, 59]. However, ride comfort should also be emphasized. On the basis of ensuring safe and efficient driving, constraints such as vehicle dynamics and evaluation index for ride comfort need to be included to achieve better optimization for generated decision.

*C. Fusion of Different Methods for Decision-Making*

Classical methods have clear levels, strong scalability and adjustability, and has the advantage of breadth traversal; while learning-based methods has concise structure, and it is suitable for processing specific scenarios with the advantage of deep traversal. The Fusion of different methods at different level of decision-making can achieve complementary advantages that should be considered in the future. For instance, the high level utilizes finite state machine method for preliminary decision, while the bottom level trains a distributed learning model based on specific scenarios for more intensive decision to realize a highly intelligent decision-making system integrating breadth and depth. In addition, how to ensure the efficient docking of different algorithms will also become the research focus of fusion methods.

*D. Robust Decision-Making in Complex Environments*

Decision-making methods need to provide instructions that are safe and efficient in complex environment. Apart from general traffic information; interaction and cooperation with other traffic participants, dynamic change of environment, weather and time variation should also be considered in the decision-making system.

TABLE V.   APPLICATIONS OF DECISION-MAKING IN EXISTING AUTONOMOUS VEHICLES

| Basic Information | | Application of Decision-Making | | | Performance |
|---|---|---|---|---|---|
| Name | Date | Methods | Inputs | Outputs | |
| CITAVT-IV [60] | 2002 | Control-based | Path generated by visual image | Longitude velocity and acceleration | Highest speed 75.6km/h on Changsha Around-city Express |
| Junior [16] | 2007 | FSM | Environmental information from sensors | High-level behaviors | Average speed 20km/h in urban environment of 2007 Darpa Challenge |
| Boss [61] | 2007 | FSM | Environmental information from sensors | High-level behaviors | Average speed 22.53km/h in urban environment of 2007 Darpa Challenge |
| Odin [18] | 2007 | FSM | Environmental information and warpoints | High-level behaviors with different priority | Average speed 20.92km/h in urban environment of 2007 Darpa Challenge |
| Talos [15] | 2007 | FSM | Drivability map with waypoints | High-level behaviors | Urban environment of 2007 Darpa Challenge. (speed not mentioned) |
| BRAiVE [62] | 2013 | FSM | Environmental information, map and driver cooperation | High-level behaviors | 13,000km in total from Italy to Shanghai. (speed not mentioned) |
| Bertha [17] | 2014 | FSM | Environmental information and digital road map | High-level behaviors with overrule of driver | Average speed 30km/h from Mannheim to Pforzheim (103km in total) |
| Intelligent Pioneer [28] | 2014 | Optimization | Environmental information and warpoints | High-level optimal behaviors | Average speed 30km/h on a long two-lanes road with two moving vehicles (2km in total) |
| Zhongtong [35] | 2019 | SVM | Three influencing factors generated from lane change scenario | Longitude velocity and steering angle | Average speed 28km/h in urban environment with decision-making accuracy of 86.27% |
| AUTOPIA [19] | 2019 | Hybrid flow diagram | Environmental information from sensors | Feasible trajectory | Longitude Speed within 10km/h to 20km/h in dynamic scenario with different driving task |

*E. Efficient End-to-End Decision-Making Method Combined with Perception System*

Decision-making methods need to be closely integrated with the perception system. It is necessary to design an end-to-end decision system with simplified architecture that adapts to more working conditions and can generate more instructions to increase the applicability of algorithms.